\documentclass[sigconf]{acmart}
\usepackage{listings}
\usepackage{multirow}

\AtBeginDocument{%
  }

\copyrightyear{2024}
\acmYear{2024}
\setcopyright{acmlicensed}\acmConference[CIKM '24]{Proceedings of the 33rd ACM International Conference on Information and Knowledge Management}{October 21--25, 2024}{Boise, ID, USA}
\acmBooktitle{Proceedings of the 33rd ACM International Conference on Information and Knowledge Management (CIKM '24), October 21--25, 2024, Boise, ID, USA}
\acmDOI{10.1145/3627673.3679122}
\acmISBN{979-8-4007-0436-9/24/10}

\begin{document}
\newcommand{\IM}{I{nfinity}M\textsc{ath}\:}
\title{\IM: A Scalable Instruction Tuning Dataset in Programmatic Mathematical Reasoning}

\author{Bo-Wen Zhang}
\authornote{The corresponding authors.}
\email{bwzhang@baai.ac.cn}
\affiliation{%
  \institution{Beijing Academy of Artificial Intelligence}
  \city{Beijing}
  \country{China}
}

\author{Yan Yan}
\authornotemark[1]
\email{yanyanustb@126.com}
\affiliation{%
  \institution{China University of Mining \& Technology Beijing}
  \city{Beijing}
  \country{China}
  }

\author{Lin Li}
\email{lilin000105@163.com}
\affiliation{%
  \institution{China University of Mining \& Technology Beijing}
  \city{Beijing}
  \country{China}
  }

\author{Guang Liu}

\email{liuguang@baai.ac.cn}
\affiliation{%
  \institution{Beijing Academy of Artificial Intelligence}
  \city{Beijing}
  \country{China}
}
\renewcommand{\shortauthors}{Bo-Wen Zhang, Yan Yan, Lin Li, \& Guang Liu}
\settopmatter{printacmref=true}

\begin{abstract}
Recent advancements in Chain-of-Thoughts (CoT) and Program-of-Thoughts (PoT) methods have greatly enhanced language models' mathematical reasoning capabilities, facilitating their integration into instruction tuning datasets with LLMs. However, existing methods for large-scale dataset creation require substantial seed data and high computational costs for data synthesis, posing significant challenges for scalability. We introduce \textbf{\IM}, a scalable instruction tuning dataset for programmatic mathematical reasoning. The construction pipeline emphasizes decoupling numbers from mathematical problems to synthesize number-independent programs, enabling efficient and flexible scaling while minimizing dependency on specific numerical values. Fine-tuning experiments with open-source language and code models, such as Llama2 and CodeLlama, demonstrate the practical benefits of \IM. These fine-tuned models, showed significant relative improvements on both in-domain and out-of-domain benchmarks, ranging from 184.7\% to 514.3\% on average. Additionally, these models exhibited high robustness on the \textit{GSM8K+} and \textit{MATH+} benchmarks, which are enhanced version of test sets with simply the number variations. \IM ensures that models are more versatile and effective across a broader range of mathematical problems. The data is available at ~\url{https://huggingface.co/datasets/flagopen/InfinityMATH}.

\end{abstract}



\maketitle


\section{Introduction}

\begin{figure}[h]
\centering
\includegraphics[width=7cm]{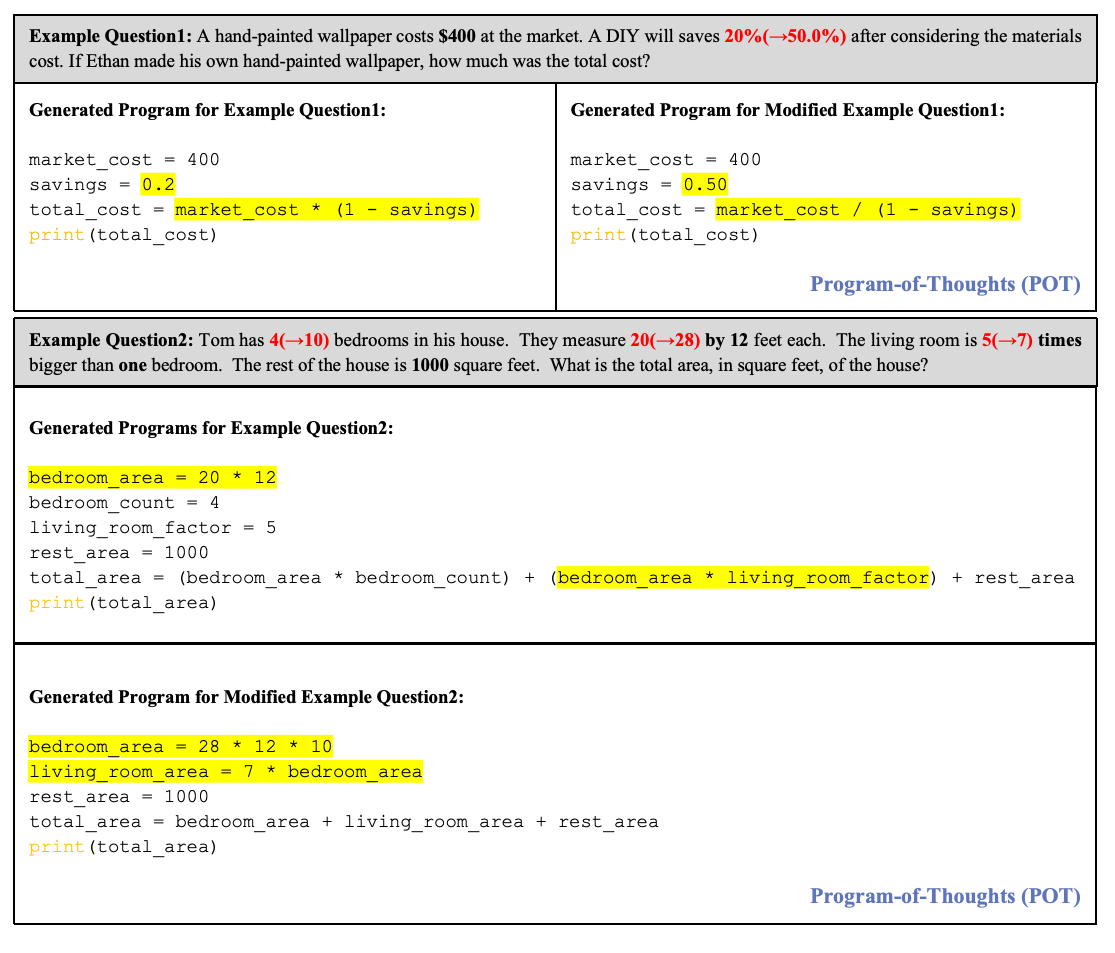}
\caption{Examples of \textit{Logical Inconsistencies in Reasoning} in LLM-generated programs when simply numerical variations}
\label{fig1:fault}
\end{figure}

Mathematical reasoning involves understanding concepts, making logical deductions, and performing complex calculations, which are essential for evaluating the overall abilities of Large Language Models (LLMs)~\cite{lample2019deep}. Enhancing model performance in mathematical reasoning is a hot research area. Several studies indicate that fine-tuning on math-specific datasets, covering problems from basic arithmetic to advanced algebra and geometry, significantly improves performance\cite{amini2019mathqa}.

Recent research highlights that CoT~\cite{wei2022chain} and PoT~\cite{chen2022program} techniques enhance mathematical reasoning capabilities through guiding models to sequentially unfold reasoning steps, or integrating executable program statements, allowing complex computations to be handled by a program interpreter. Therefore, several studies generate reasoning processes or programming solutions for mathematical problems using models like GPT-4 to synthesize instruction tuning datasets, like OpenMathInstruct-1~\cite{toshniwal2024openmathinstruct}, MetaMathQA~\cite{yu2023metamath} and MathInstruct~\cite{yue2023mammoth}. However, the limited availability of high-quality mathematical reasoning problems and the high computational cost for data synthesis restrict large-scale data production.

In the context of these scaling challenges, we observed \textit{Logical Inconsistencies in Reasoning} in LLM-generated programs. Figure~\ref{fig1:fault} shows that minor numerical variations in problems lead to unexpected changes in the program's calculation logic, resulting in reasoning errors. For instance, in \textbf{Example Question 1}, altering a discount from 20\% to 50\% results in an illogical switch from multiplication to division, contradicting consistent discount principles. Similarly \textbf{Example Question 2} highlights a spatial reasoning error where changing the bedroom size results in an incorrect adjustment to the calculation logic. These examples show that slight numerical changes disrupt the program's logic, revealing issues with the reasoning robustness to numerical variations.

We propose \textbf{\IM}, a scalable instruction tuning dataset for programmatic mathematical reasoning. The construction of \IM involves a multi-step pipeline: numerical values in mathematical problems are firstly identified and abstracted to generate "universal templates" for questions, then an LLM (such as GPT-4) generates programs independent of these specific values, and finally, the templates are repopulated with varied numbers to broaden the dataset while preserving reasoning logic. Consequently, \IM comprises 101,380 extensible data points generated from 7 high-quality mathematical datasets, formulating an \textbf{"infinite mathematical instruction tuning dataset"}.

We fine-tuned the 7B versions of the Llama2 and Aquila2 (language models), as well as the Codellama (code foundation model), with the \IM dataset and evaluated on four in-domain and five out-of-domain benchmarks. Empirical results indicate these fine-tuned models consistently outperform other state-of-the-art models of comparable size on the most benchmarks. Furthermore, to investigate the \textit{Logical Inconsistencies in Reasoning} phenomenon, we created enhanced versions of the GSM8K and MATH test sets, designated GSM8K+ and MATH+, which only modified the numerical values within the problems. Comparative analysis reveals that models trained on \IM exhibit superior accuracy and robustness compared to those trained on other PoT datasets.

\section{Related Work}

The use of large models for solving mathematical problems has become a key research focus, serving as a crucial indicator for assessing the performance of LLMs in complex multi-hop and quantitative reasoning tasks. Researchers have explored various methods to enhance the mathematical reasoning capabilities of LLMs, bridging the gap between closed-source and open-source models.


The Chain-of-Thought (CoT) method, introduced by Wei et al.\cite{wei2022chain}, decomposes mathematical problems into smaller, interconnected tasks. This step-by-step approach enables models to solve complex issues incrementally. Wang et al.\cite{wang2022self} enhanced this with the Self-Consistency method, where the model generates multiple reasoning processes and selects the most likely correct answer through voting. Li et al.~\cite{li2022making} further developed this by transforming a single prompt into multiple ones, checking intermediate reasoning steps, and applying weighted voting. However, CoT requires language models to generate and compute mathematical expressions, which can be inefficient for tasks like solving polynomial equations or calculus problems.


Chen et al.\cite{chen2022program} introduced Program-of-Thoughts (PoT), which delegates computational steps to an external interpreter (e.g., Python), while maintaining natural language reasoning. Luo et al.\cite{luo2023wizardcoder}, using code-specific Evol-Instruct, fine-tuned StarCoder~\cite{li2023starcoder} to create WizardCoder, outperforming major LLMs on code generation benchmarks. Bi et al.\cite{bi2024program} proposed the Code and Instruction Reasoning Score (CIRS), evaluating the correlation between code and reasoning abilities. Their experiments on datasets like AsDiv\cite{miao2021diverse} and GSM8K~\cite{cobbe2021gsm8k} highlighted PoT's superiority over CoT, especially in handling complex computations with external libraries like SymPy.


Fine-tuning methods leverage the generic features learned by large models to perform well on new tasks with minimal training. Yue et al.\cite{yue2023mammoth} introduced MathInstruct, a dataset mixing CoT and PoT principles, fine-tuned on Llama22\cite{touvron2023llama} and codellama~\cite{roziere2023code}. Yu et al.\cite{yu2023metamath} proposed MetaMathQA, based on GSM8K and MATH, incorporating more reasoning paths and fine-tuned on LLaMA-2. Toshniwal et al.\cite{toshniwal2024openmathinstruct} introduced OpenMathInstruct-1, synthesizing solutions from GSM8K and MATH's code interpreters, fine-tuned on Mistral~\cite{jiang2024mixtral}, Llama-2, and CodeLlama.

\section{Methodology}
The motivation for constructing \IM is to tackle the challenge of scaling data while addressing logical inconsistencies in reasoning. We propose a simple yet efficient pipeline that decouples numerical values from mathematical problems and synthesizes a large number of similar problems without significantly increasing computational costs. The goal is to overcome the dependencies of problem synthesis on specific numerical values, thereby maximizing data utilization and enhancing the robustness of models.

\begin{figure}[h]
\centering
\includegraphics[width=8cm]{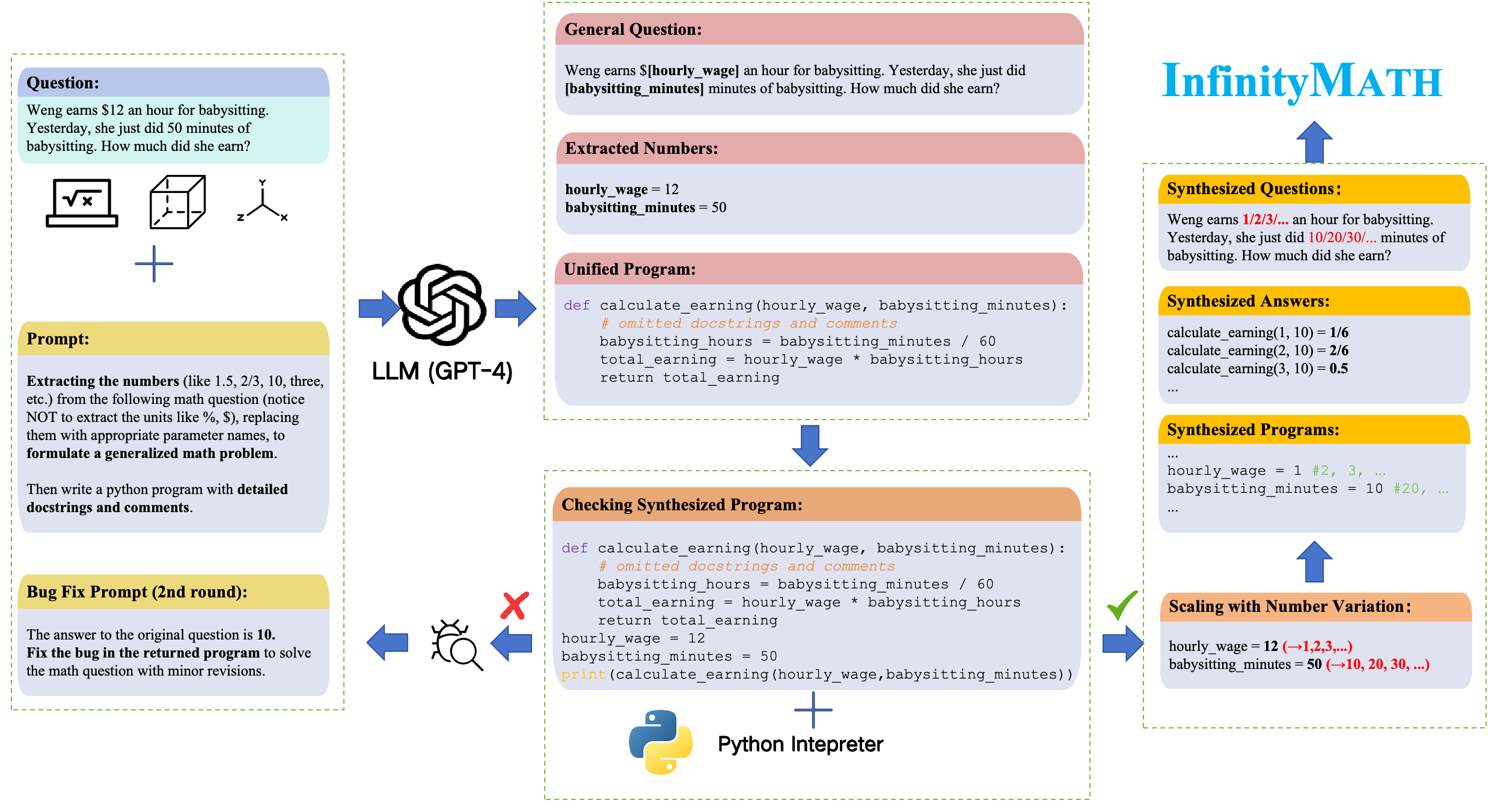}
\caption{The construction pipeline for \IM}
\label{fig3:sprint}
\end{figure}

\begin{table*}[h]
    \centering
    \caption{Statistics of \IM, the number of questions in source datasets and successful generated samples.}
    \resizebox{\textwidth}{!}{%
    \begin{tabular}{|c|c|c|c|c|c|c|c|c|}
        \hline
        \textbf{Dataset} & \textbf{AQuA-RAT~\cite{ling2017program}} & \textbf{GSM8K~\cite{cobbe2021gsm8k}} & \textbf{MATH~\cite{hendrycks2021measuring}} & \textbf{NUMGLUE~\cite{mishra2022numglue}} & \textbf{MathQA~\cite{amini2019mathqa}} & \textbf{TheoremQA~\cite{chen2023theoremqa}} & \textbf{Deepmind-Mathematics~\cite{saxton2019analysing}} & \textbf{Total} \\ \hline
        \textbf{Samples} & 53822 & 7254 & 5021 & 14228 & 20319 & 310 & 426 & 101380 \\ \hline
        \textbf{No. of Question} & 97467 & 7473 & 7500 & 20359 & 29837 & 800 & 490 & - \\ \hline
        \textbf{Success Rate} & 55.23\% & 97.07\% & 66.95\% & 69.89\% & 68.12\% & 38.75\% & 86.94\% & - \\ \hline
    \end{tabular}%
    }
    
    \label{tab:infinitymath}
\end{table*}
\subsection{Data Synthesis}
Drawing from algebraic thinking, we hypothesize that each mathematical problem can be transformed into a more general, number-independent ``generic problem" and ``generic solution". Therefore, when using LLMs (such as GPT-4~\cite{achiam2023gpt}) for data synthesis, we diverge from the classical PoT method by not directly generating solution code. Instead, we employ a multi-step process:

First, numerical constants in problems are masked to create ``generic problems", with these masked parts replaced by appropriate variable names as placeholders. This generalization simplifies program synthesis and prepares for scalability. Subsequently, LLMs are utilized to generate programs that solve these generic problems. Specifically, the generation of function call-based programs instead of inline code is required. This approach enhances the reusability of the synthesized code and supports efficient scaling. Moreover, including a docstring in the function code improves readability by describing the function's purpose, the variables used, and the expected output. This establishes a clear relationship between the solution and the variables. Comments within the program explain the computational logic of each line, incorporating CoT rationale-like content. Therefore, the synthesized programs must include docstrings and comments to ensure clarity and comprehensibility.

To minimize LLM usage and reduce computational costs, we designed and compared several versions of prompt templates. Ultimately, we developed a multi-task prompt that requires the LLM to complete the tasks described above simultaneously, as illustrated in Figure~\ref{fig3:sprint}. Moreover, we included an example to leverage the in-context learning capability of the LLM, enhancing the model's accuracy. Additionally, to facilitate post-processing, we strictly defined the format of the returned answers, thereby increasing the usability of the synthesized data.

Afterward, using simple rules, the program solution to the original math problem can be synthesized directly from the \textbf{``General Question"}, \textbf{``Extracted Numbers"} and \textbf{``Unified Program"} obtained in the 1st-round response. The accuracy of the data synthesis process is verified by executing the synthesized code with a Python interpreter and comparing the results to the ground truth answer. For more complex mathematical problems, the generated programs often contain bugs. Instead of repeatedly adjusting hyperparameters inefficiently, we leverage the multi-turn dialogue capability of LLMs. By providing feedback from the checking program, we construct a \textbf{Bug Fix Prompt} for the 2nd round, supplying the LLM with the correct answer to the original problem and requesting minimal modifications to the erroneous code, as shown in Figure~\ref{fig3:sprint}.

To enhance the model's generalization capability in mathematical reasoning, we selected seven high-quality open-source math datasets and used the aforementioned method to generate solutions. The statistics of \IM (each problem corresponding to one data point) is shown in Table~\ref{tab:infinitymath}.

\subsection{Scaling with Data Augmentation}
Recent data augmentation studies have focused on using LLMs to rewrite problems or generate different solutions for the same problem. However, there has been limited research on approaches that do not rely on additional LLM usage. To address this gap, we conducted the following work.

During data synthesis, we generated generic mathematical problems and function call-based programs. As shown in Figure~\ref{fig3}, new problems and solution code can be generated by reversing the process: replacing variable placeholders with numbers.

Suppose the original mathematical problem uses $k$ numbers, which are replaced with variable placeholders during data synthesis. Since we can choose whether to replace variables with numbers or to retain the variable placeholder, there are $2^k-1$ replacement options for each group of reasonable variable-to-number mappings, resulting in $2^k-1$ possible program solutions. 

When modifying the program, we remove variable assignments and docstring parts related to the replaced variables, then replace occurrences of the variables with the original numbers. We verify correctness by checking if the modified code runs correctly and produces the expected output, ensuring the generated programs are logically correct.

It is crucial that replacement numbers follow certain rules to maintain the intended meaning (e.g., ensuring the number of people is an integer). Therefore, we require the LLM to provide reasonable number ranges or criteria during synthesis to ensure the validity of the numbers.

\section{Experiments}

\begin{table*}[htbp]
\footnotesize
\caption{The overall evaluation results on both in-domain and out-of-domain benchmarks.}
\centering
\begin{tabular}{|c|c|c|c|c|c|c|c|c|c|c|}
\hline
\multirow{2}{*}{Base Model} & \multirow{2}{*}{Training Dataset} & \multicolumn{5}{c|}{In-domain Evaluation} & \multicolumn{4}{c|}{Out-of-domain Evaluation} \\
\cline{3-11}
 &  & GSM8K & MATH & AQuA & Mathematics & NumGLUE & SVAMP & SimulEq & SAT-Math & MMLU-Math \\
\hline
\multirow{6}{*}{Llama2} & - & 14.60 & 2.50 & 30.30 & 6.20 & 29.90 & 34.50 & 4.60 & 22.70 & 30.60 \\
 & WizardMath & 54.90 & 10.70 & 26.30 & 9.30 & 36.10 & 36.10 & 12.80 & 25.40 & 31.10 \\
 & MetaMathQA & \textbf{66.50} & 19.80 & - & - & - & - & - & - & - \\
 & MathInstruct & 53.60 & \textbf{31.50} & \textbf{44.50} & 46.30 & 61.20 & 67.70 & 41.20 & \textbf{42.70} & \textbf{42.60} \\
 & MathInstruct(pot) & 47.23 & 24.44 & 15.75 & 41.00 & 63.82 & 59.40 & 39.88 & 10.91 & 20.53 \\
 & InfinityMath & 60.80 & 29.40 & 38.58 & \textbf{63.70} & \textbf{75.91} & \textbf{76.10} & \textbf{56.03} & 25.91 & 30.70 \\
 & & (+316.44\%) & (+1076.00\%) & (+27.30\%) & (+927.42\%) & (+153.88\%) & (+120.58\%) & (+1118.09\%) & (+14.13\%) & (+0.33\%) \\
\hline
\multirow{4}{*}{CodeLlama} & - & 16.30 & 10.92 & 18.50 & 28.60 & 26.68 & 50.60 & 3.89 & 15.91 & 20.77 \\
 & MathInstruct & 59.40 & 33.40 & \textbf{47.20} & 55.40 & 66.40 & 71.40 & 45.90 & \textbf{40.50} & \textbf{48.30} \\
 & MathInstruct(pot) & 56.86 & 29.88 & 16.54 & 61.70 & 62.19 & 69.60 & 40.27 & 11.82 & 17.66 \\
 & InfinityMath & \textbf{65.80} & \textbf{34.06} & 38.18 & \textbf{67.10} & \textbf{71.40} & \textbf{77.00} & \textbf{74.71} & 32.73 & 37.27 \\
 & & (+303.07\%) & (+211.91\%) & (+106.38\%) & (+134.65\%) & (+167.62\%) & (+52.17\%) & (+1820.05\%) & (+105.76\%) & (+79.47\%) \\
\hline
\multirow{4}{*}{Aquila2} & - & 20.47 & 3.78 & 14.96 & 4.80 & 13.82 & 23.30 & 6.23 & 19.09 & 12.63 \\
 & MathInstruct & 50.19 & 25.98 & 15.35 & \textbf{71.79} & 61.30 & 44.40 & 40.08 & 14.55 & 17.86 \\
 & MathInstruct(pot) & 49.66 & \textbf{26.50} & 12.60 & 47.40 & \textbf{72.36} & 50.10 & 46.11 & 17.27 & \textbf{26.80} \\
 & InfinityMath & \textbf{51.25} & 26.18 & \textbf{23.62} & 42.20 & 68.04 & \textbf{66.00} & \textbf{59.92} & \textbf{19.55} & 21.77 \\
 & & (+150.40\%) & (+592.59\%) & (+57.88\%) & (+779.17\%) & (+392.12\%) & (+183.26\%) & (+861.97\%) & (+213.82\%) & (+72.41\%) \\
\hline
\end{tabular}
\end{table*}

\subsection{Experimental Setup}

We selected the open-source models CodeLlama, Llama2, and Aquila2, each with 7B parameters. These models were fine-tuned using \IM and other datasets to validate instruction tuning. CodeLlama and Llama2 were aligned to an Alpaca-like instruction structure, while Aquila2 followed the Aquila-v2 structure. We used a learning rate of $2 \times 10^{-5}$ for CodeLlama and Llama2, and $1 \times 10^{-5}$ for Aquila2, with a global batch size of 128 for all models.

We evaluated the effectiveness of our data on five in-domain test sets: GSM8K, MATH, AQuA-RAT, NumGLUE, Mathematics, and four out-of-domain test sets: SVAMP~\cite{patel2021nlp}, SimulEq~\cite{koncel2016mawps}, SAT-Math~\cite{zhong2023agieval}, MMLU-Math~\cite{hendrycks2020measuring}. For datasets containing both CoT and PoT, we used the Mammoth~\cite{yue2023mammoth} evaluation framework, which involves first evaluating with a PoT prompt. If the generated program fails, a second evaluation is done using a CoT prompt to potentially improve results. For datasets with only PoT prompts (including \IM), we exclusively used PoT prompts for evaluation with no retries. All evaluations were performed in a \textbf{0-shot setting} without additional examples.

\subsection{Experimental Results}

\subsubsection{Main Results}
The evaluation results in Table 2 show that the \IM dataset consistently enhances performance across different base models compared to other datasets.

For in-domain evaluation, \IM demonstrates substantial improvements. For example, with \IM, Llama2 achieves a 316.44\% improvement on GSM8K, 1076.00\% on MATH, and 927.42\% on Mathematics. Similar gains are observed with CodeLlama and Aquila2. In out-of-domain evaluation, \IM continues to show significant performance gains. With \IM, Llama2 achieves a 120.58\% improvement on SVAMP and 1118.09\% on SimulEq. Notable improvements are also seen with CodeLlama and Aquila2.

While \IM does not perform as well as MathInstruct on some benchmarks due to the inclusion of CoT data and the advantage of two evaluation opportunities, it still consistently enhances performance across all evaluated benchmarks. This highlights the effectiveness and robustness of the \IM dataset in improving the mathematical reasoning capabilities of LLMs.

\subsubsection{Effect of Scaling}
We used subsets of data from GSM8K and MATH within \IM to compare the effectiveness of scaling methods. By replacing the variables in the \IM subsets with different numbers, we fine-tuned CodeLlama and Llama2 models. To ensure that the improvements were due to the scaling method and not merely the increased data volume, we compared checkpoints based on the same number of training steps rather than epochs. The results, shown in Table~\ref{tab:aug}, demonstrate that augmenting the GSM8K and MATH subsets of InfinityMath with different numbers increased data volume and improved accuracy. This validates the effectiveness of our synthetic dataset method.

\begin{table}[htbp]
\footnotesize
\caption{The results of fine-tuning using data augmentation with \IM subsets on GSM8K and MATH~(G+M).}
\centering
\begin{tabular}{|c|c|c|c|}
\hline
BaseModel & \IM-G+M & GSM8K & MATH \\
\hline
\multirow{2}{*}{Llama2} & w/o scaling & 51.78 & 22.78 \\
 & w/ scaling & 55.42 (+7.03\%) & 23.98 (+5.27\%) \\
\hline
\multirow{2}{*}{CodeLlama} & w/o scaling & 62.02 & 30.86 \\
 & w/ scaling & 63.38 (+2.19\%) & 31.32 (+1.49\%) \\
\hline
\end{tabular}
\label{tab:aug}
\end{table}

\subsubsection{Results of GSM8K+ and MATH+}
\vspace{-0.5cm} 
\begin{table}[htbp]
\footnotesize
\caption{The comparison results on GSM8K+ and MATH+.}
\centering
\begin{tabular}{|c|c|c|c|c|c|c|c|}
\hline
\multirow{2}{*}{Base Model} & \multirow{2}{*}{Training Dataset} & \multicolumn{3}{|c|}{GSM8K+} & \multicolumn{3}{|c|}{MATH+} \\
\cline{3-8}
 &  & x & y & \% & x & y & \% \\
\hline
\multirow{2}{*}{Llama2} 
 & MathInstruct(pot) & 827 & 463 & 56.0 & 1484 & 581 & 39.1 \\
 & InfinityMath & \textbf{974} & \textbf{605} & \textbf{62.1} & \textbf{1741} & \textbf{684} & \textbf{39.3} \\
\hline
\multirow{2}{*}{Codellama} 
 & MathInstruct(pot) & 913 & 568 & 62.2 & 1720 & 696 & 40.5 \\
 & InfinityMath & \textbf{1022} & \textbf{672} & \textbf{65.8} & \textbf{1985} & \textbf{850} & \textbf{42.8} \\
\hline
\end{tabular}
\label{tab:evalplus}
\end{table}
To investigate logical inconsistencies in reasoning within LLM-generated programs, we constructed the GSM8K+~\footnote{Available at \url{https://huggingface.co/datasets/flagopen/}} and MATH+~\footnote{Available at \url{https://huggingface.co/datasets/flagopen/MATHplus}} evaluation datasets. By replacing two sets of numbers in the original datasets and conducting rigorous manual checks, we created $1319 \times 3$ and $3818 \times 3$ evaluation data points, respectively. The evaluation results are shown in Table~\ref{tab:evalplus}, where $x$ and $y$ represent the number of instances where the model generated at least one correct answer and all three correct answers for the same problem, respectively. The $y/x$ ratio indicates the preservation of logical consistency in reasoning.



\subsubsection{DocString Ablation}

To analyze the impact of CoT rationale-like descriptions, we conducted an ablation study focusing on docstrings. By removing docstrings in \IM, we performed comparative experiments using CodeLlama on four in-domain evaluation sets. The experimental results, shown in Table~\ref{tab:ablation}, indicate that even though docstrings do not directly affect program execution, they enhance the model's reasoning capabilities.

\begin{table}[htbp]
\footnotesize
\caption{Impact of docstrings on the 7B codellama model.}
\centering
\begin{tabular}{|c|c|c|c|c|}
\hline
 & AQuA & GSM8K & MATH & NUMGLUE \\
\hline
w/o docstring (\%) & 31.5 & 58.7 & 28.2 & 69.9 \\
w/ docstring (\%) & 39.4 & 62.3 & 31.1 & 68.0 \\
\hline
\end{tabular}
\label{tab:ablation}
\end{table}

\section{Conclusion}
We open-source \IM, a mathematical reasoning instruction tuning dataset with each data point including a generic problem and a solution template, allowing for easy scaling into an infinite dataset. The fine-tuning experiments show that \IM effectively alleviates logical inconsistencies in reasoning.
\section*{Acknowledgements}
This work was supported by National Key R\&D Program of China (2022ZD0116312).

\bibliographystyle{ACM-Reference-Format}
\balance
\bibliography{sample-base}

\end{document}